\documentclass[lettersize,journal]{IEEEtran}
\usepackage{amsmath,amsfonts}
\usepackage{algorithmic}
\usepackage{array}
\usepackage[caption=false,font=normalsize,labelfont=sf,textfont=sf]{subfig}
\usepackage{textcomp}
\usepackage{stfloats}
\usepackage{url}
\usepackage{verbatim}
\usepackage{graphicx}
\usepackage{graphics}
\usepackage{epsfig}
\usepackage{caption}
\usepackage{amssymb}  
\usepackage{makecell}
\usepackage{booktabs}
\usepackage{multirow}
\usepackage{threeparttable} 
\usepackage{algorithm}
\usepackage{algorithmic}
\def\BibTeX{{\rm B\kern-.05em{\sc i\kern-.025em b}\kern-.08em
    T\kern-.1667em\lower.7ex\hbox{E}\kern-.125emX}}
\usepackage{balance}
\begin{document}
\title{Mesh-LOAM: Real-time Mesh-Based LiDAR Odometry and Mapping 
}

\author{Yanjin Zhu, Xin Zheng, and Jianke Zhu
~\IEEEmembership{Senior Member,~IEEE
}
\thanks{All authors are with the College of Computer Science and Technology, Zhejiang University, 38 Zheda Road, Hangzhou, China. } %
\thanks{ E-mail: \tt\small \{yanjinzhu,xinzheng,jkzhu\}@zju.edu.cn} %
\thanks{Jianke Zhu is the Corresponding Author.}%
}

\markboth{Journal of \LaTeX\ Class Files,~Vol.~18, No.~9, September~2020}%
{How to Use the IEEEtran \LaTeX \ Templates}

\maketitle

\begin{abstract}
Despite having achieved real-time performance in mesh construction, most of the current LiDAR odometry and meshing methods may struggle to deal with complex scenes due to relying on explicit meshing schemes. They are usually sensitive to noise. To overcome these limitations, we propose a real-time mesh-based LiDAR odometry and mapping approach for large-scale scenes via implicit reconstruction and a parallel spatial-hashing scheme.
To efficiently reconstruct triangular meshes, we suggest an incremental voxel meshing method that updates every scan by traversing each point once and compresses space via a scalable partition module. By taking advantage of rapid accessing triangular meshes at any time, we design point-to-mesh odometry with location and feature-based data association to estimate the poses between the incoming point clouds and the recovered triangular meshes. The experimental results on four datasets demonstrate the effectiveness of our proposed approach in generating accurate motion trajectories and environmental mesh maps.

\end{abstract}


\section{Introduction}
LiDAR odometry and mapping (LOAM) aims to estimate the 6-DoF poses of moving sensors while continuously constructing the surrounding environment, which serves as a fundamental component in robotic applications~\cite{chalvatzaras2022survey, bresson2017simultaneous}. There is demand for a representative and compact map structure, that can not only improve the real-time capability of odometry but also allow for seamless integration with downstream tasks. However, prevalent map representations such as, point cloud~\cite{zhang2014loam, wang2021f, vizzo2023kiss}, grid map~\cite{hornung2013octomap, zhou2023lidar} and surfel~\cite{behley2018suma, chen2019suma++} cannot fulfill these requirements. Specifically, it requires substantial amounts of memory resources to store point clouds in large-scale scenes. Although grid maps and surfels have more compact map representations, they suffer from a lack of connectivity between their elemental units. Such discontinuity presents substantial challenges for direct robotic utilization, necessitating complex and resource-intensive post-processing techniques~\cite{piazza2018real}.   

   \begin{figure}[thpb]
      \centering
      \includegraphics[width=0.95\linewidth]{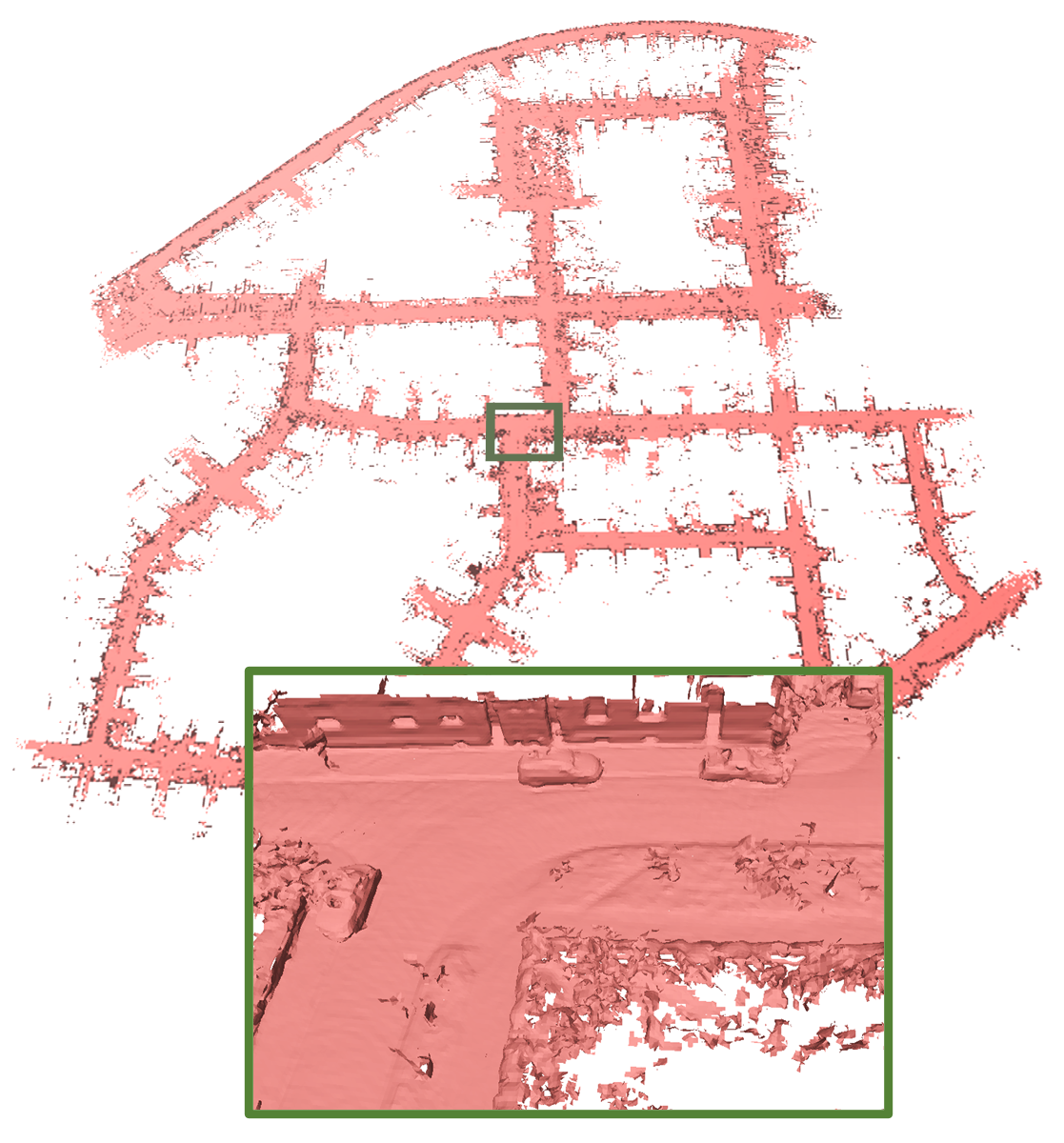}
      \caption{Our odometry and mapping result on sequence `00' of the KITTI odometry dataset~\cite{geiger2012kitii}. Our proposed Mesh-LOAM can accurately estimate the poses of a moving vehicle while simultaneously obtaining a precise and compact mesh of the large-scale outdoor scene. The whole algorithm runs at around 54 frames per second.  }
      \label{fig:kitti00}
   \end{figure}

In comparison, the triangular mesh using a collection of vertices and triangular facets offers a concise and accurate depiction of outdoor environments. It is widely adopted in dense map generation~\cite{kuhner2020large}, pose estimation~\cite{vizzo2021puma, slamesh}, loop closure~\cite{schops2019surfelmeshing}, and collision detection~\cite{ericson2004real}. Mesh maps provide a smooth continuous surface representation, which better supports robot navigation and decision making in the environment. In spite of some works~\cite{vizzo2021puma,slamesh} focusing on mesh-based LOAM, it still remains a challenging problem to perform real-time meshing in large-scale environments. For instance, implicit reconstruction like Poisson surface reconstruction~\cite{vizzo2021puma} is utilized to construct detailed and accurate smooth surface meshes of outdoor scenes, whose performance is far from real-time. Recent studies~\cite{slamesh, lin2023immesh} make use of either efficient hierarchical sparse data structures or techniques to realize real-time meshing. They utilize explicit methods like Delaunay triangulation~\cite{lin2023immesh} or the Gaussian Process~\cite{slamesh} to incrementally construct a mesh map. However, these explicit approaches are sensitive to noise, which require high-quality input data. Moreover, they suffer from increasing complexity when the target scenario has a complicated structure deviating from the Manhattan assumptions. 

To address the above limitations, we propose a real-time large-scale mesh-based LiDAR odometry and mapping approach named Mesh-LOAM. In this paper, we aim to improve the geometrical accuracy of LiDAR-based mapping and simultaneously reduce the drift of estimated poses. Moreover, we adopt the implicit reconstruction by using an implicit function that leverages the advantages of its robustness to noise and ability to handle complex scenes. In order to achieve the real-time implicit reconstruction for large-scale scenes, we propose an incremental voxel meshing method under a parallel spatial-hashing scheme, where our passive computational model for SDF value and scalable partition module are able to accelerate the computation. Furthermore, point-to-mesh odometry is employed to estimate the poses between the incoming point clouds and the reconstructed triangular meshes. Fig.~\ref{fig:kitti00} shows an example of our Mesh-LOAM output with a triangular mesh map on the KITTI odometry dataset.

The main contributions of this paper are: 1) a real-time mesh-based LiDAR odometry and mapping method for large-scale scenes using a parallel spatial-hashing scheme;
2) an incremental voxel meshing approach integrates each LiDAR scan with only one traversal, which takes advantage of a scalable partition module; 3) an accurate point-to-mesh odometry method to estimate pose from point clouds and reconstructed triangular meshes; 4) experiments show that our proposed Mesh-LOAM approach achieves high accuracy of estimated poses and simultaneously recovers promising triangular mesh in real-time for large-scale outdoor scenes.

\section{RELATED WORK}
\subsection{Point-based LiDAR Odometry and Mapping}
Point-based LOAM refers to the utilization of point clouds as maps or for localization in LiDAR-based systems.
The pioneering work in LiDAR odometry and mapping has been achieved by LOAM~\cite{zhang2014loam}. They extract edge features and plane features, which are used to calculate point-to-line and point-to-plane distances to a grid-based voxel.
An optimized registration LiDAR SLAM framework is FLOAM~\cite{wang2021f}, which achieves competitive localization accuracy at a low computational cost and runs at more than 20 Hz.
Deschaud~\cite{deschaud2018imls} proposes IMLS-SLAM, where scan-to-model matching with Implicit Moving Least Squares (IMLS) surface representation improves the odometry accuracy. Unfortunately, IMLS-SLAM requires intensive computation. Zheng and Zhu~\cite{zheng2021elo} propose a novel efficient GPU-based LiDAR odometry approach, which takes advantage of both non-ground spherical range image and ground bird’s-eye-view map. Behley and Stachniss~\cite{behley2018suma} use a surfel-based map to represent the large-scale environment and perform projective data association of the current scan to the surfel maps. Each surfel contains a position, a normal, and other attributes. Zhou et al.~\cite{zhou2023hpplo} propose an unsupervised method to estimate 6-DoF poses using a differentiable point-to-plane solver. These point-to-plane ICP and its variants take account of the normal vector information on the surface of the point cloud, which usually provide accurate odometry results.

Recently, KISS-ICP~\cite{vizzo2023kiss} achieves encouraging performance using a robust point-to-point ICP. Lin et al.~\cite{lin2023immesh} propose a novel meshing framework for simultaneous localization and meshing in real-time, where their localization still relies on point cloud maps. 

\subsection{Mesh-based LiDAR Odometry and Mapping}
This paper emphasizes mesh-based LiDAR odometry and mapping methods, which utilize meshes to not only depict the surrounding environment but also facilitate localization.
Vizzo et al.~\cite{vizzo2021puma} represents the map as triangle meshes computed via Poisson surface reconstruction and performs point-to-mesh ICP. This work demonstrates that triangular meshes are well-suited for the registration of 3D LiDAR scans. However, it requires saving LiDAR scans in a sliding windows fashion and repeating multiple calculations. This results in a significant overhead on computational time. To achieve a real-time LiDAR-based simultaneous localization and meshing system, Ruan et al.~\cite{slamesh} employ the Gaussian process reconstruction to reconstruct meshes by adopting domain decomposition and local regression techniques~\cite{ruan2020gp}. Although SLAMesh can simultaneously build a mesh map and perform localization against the mesh map, it becomes significantly complicated in modeling complicated geometries. Compared to point-based LOAM methods, mesh-based LOAM has fewer research studies. In addition, the accuracy of these point-to-mesh methods is not so good as some approaches~\cite{vizzo2023kiss, wang2021f}, which directly use point cloud maps by point-to-point or point-to-plane constraints. 

   \begin{figure*}[thpb]
      \centering
      \includegraphics[width=\linewidth]{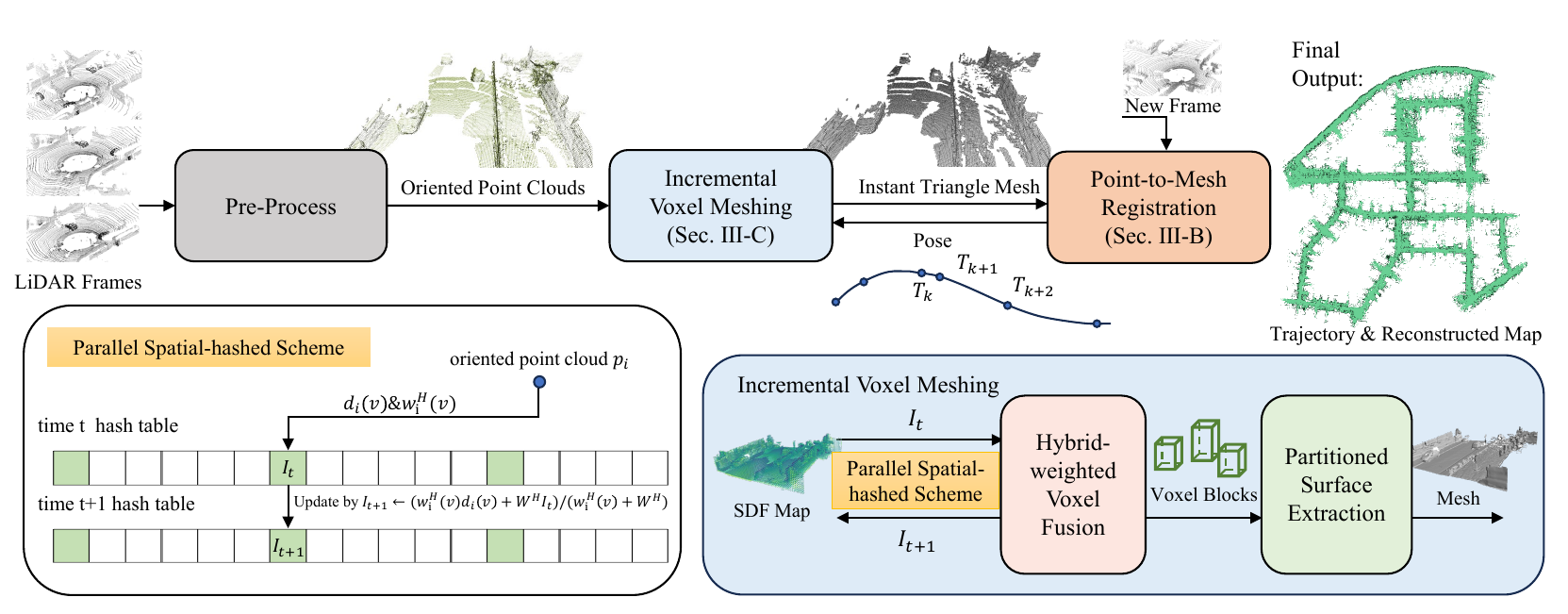}
      \caption{Overview of our proposed Mesh-LOAM. 
      Firstly, oriented point clouds are obtained by using Principal Component Analysis (PCA) to estimate normals. Secondly, we use incremental voxel meshing to obtain instant triangular mesh. Finally, point-to-mesh odometry is performed for the new LiDAR frames and the reconstructed triangular meshes to estimate the pose $\mathbf{T}_k$. 
      The hybrid-weighted voxel fusion, based on the parallel spatial-hashing scheme, updates the SDF value $I_{t+1}$ of any oriented point cloud $p_i$ by considering the distance $d_i(v)$ and the previous SDF value $I_t$ in the corresponding voxel. The parallel retrieval for height-adaptive voxel blocks allows partitioned surface extraction.
      }
      \label{fig:overview}
   \end{figure*}

\subsection{Offline LiDAR Reconstruction}
Various kinds of implicit functions, i.e., radial basis function and its variants~\cite{carr2001rbf, kolluri2008provably}, tangent plane~\cite{hoppe1992surface}, signed distance function~\cite{ortiz2022isdf}, truncated signed distance function~\cite{curless1996volumetric, vizzo2022vdbfusion}, are utilized to approximate the underlying surface. 
These implicit methods have proven to be computationally intensive or constrained by small scale, which hinders their deployment in real-time applications with a large number of LiDAR scans. The Implicit Moving Least Squares (IMLS), known for its excellent approximation to the signed distance function (SDF) of the surface~\cite{kolluri2008provably}, is quite time-consuming in the case of large-scale outdoor scenes. 

Alternatively, voxel-based methods make use of the voxel structure to retain the completeness of point set information. 
Curless and Levoy~\cite{curless1996volumetric} first incrementally construct a TSDF localized near the input points. 
To fulfill large-scale reconstruction, Fuhrmann et al.~\cite{fuhrmann2011fusion} extend~\cite{curless1996volumetric} through constructing a hierarchical signed distance field. However, it takes considerable time to obtain the reconstruction results.
While K{\"u}hner et al.~\cite{kuhner2020large} present a volumetric depth fusion for large-scale mapping, they require multiple passes over the same scene for compelling reconstruction results.

Learning-based methods~\cite{zhong2022shine, ortiz2022isdf, deng2023nerf} usually employ a shallow neural network to represent large-scale 3D scenes. Despite achieving encouraging mapping results, these methods still exhibit several limitations. For instance, they are hard to achieve real-time performance and
rely on prior knowledge of the target scene's volume. In addition, the size of the bounding box needs to be appropriately set. Our proposed approach can deal with scenes of unknown volume without a predetermined bounding box while fulfilling real-time requirements.

\subsection{Online LiDAR Reconstruction}
Voxblox~\cite{oleynikova2017voxblox} allows dynamically growing maps by employing voxel hashing~\cite{niessner2013voxelhashing} as the underlying data structure and incrementally estimates Euclidean Signed Distance Fields (ESDFs) from TSDFs. However, it may perform inferior in reconstructing large-scale scenes. VDBFusion~\cite{vizzo2022vdbfusion} is a CPU-based flexible and effective mapping system. Niedzwiedzki et al.~\cite{IDTMM} introduce a new incremental triangular mesh generation algorithm. Nevertheless, they both rely on other odometry methods to provide pose information.

\section{Mesh-based LiDAR Odometry and Mapping}
\subsection{Overview}
In this section, we present a real-time approach to large-scale mesh-based LiDAR odometry and mapping. Fig.~\ref{fig:overview} shows the overview of our proposed pipeline. 
Firstly, we present point-to-mesh odometry with a location and feature-based data association module to estimate poses. Secondly, we propose an efficient voxel meshing method that takes advantage of sparse voxels to incrementally reconstruct surface meshes.
Finally, we introduce a simple and effective parallel spatial hash-based scheme and implementation to efficiently retrieve the voxel and guarantee the sustainability of reconstruction. 

\subsection{Point-to-Mesh LiDAR Odometry}
We adopt a point-to-mesh registration similar to Puma~\cite{vizzo2021puma} and SLAMesh~\cite{slamesh}, which can be utilized to improve the accuracy of the odometry. The management of triangular meshes details in Section~\ref{Incremental Voxel Meshing} and Section~\ref{voxelhashing}. Since scan-to-model matching performs better than classical scan-to-scan matching~\cite{zheng2021elo, behley2018suma, chen2019suma++}, our mesh representations are computed from consecutively accumulated scans.   

\subsubsection{Planar Feature Selection}
In the proposed point-to-mesh registration framework, we select planar points to facilitate accurate pose estimation. The process involves with estimating the surface normal of a 3D point $\mathbf{p} = (x,y,z)^{\top}$ by fitting a local plane within the current LiDAR scan. In this context, we assume a plane in 3D Euclidean space is characterized by a normal vector $\mathbf{n} = (n_x,n_y,n_z)^{\top}$ and a scalar value $d$, satisfying the condition  $\mathbf{p}^{\top} \mathbf{n} = d$. The computation of the normal vector for each point is achieved by solving the following equation:
\begin{equation}
    e = \sum\limits_{i=1}^{k} (\mathbf{p}_i^{\top}\mathbf{n}-d)^2 \quad s.t. \quad \mathbf{n}^{\top}\mathbf{n}=1  .
    \label{equ:normal}
\end{equation}
Minimizing the above equation can be converted into performing principal component analysis on the covariance matrix constructed from the neighborhoods of each point $\mathbf{p}_i$. The covariance matrix $C$ is computed by
\begin{equation}
    C=\frac{1}{k} \sum_{i=1}^{k} (\mathbf{p}_i- \Bar{\mathbf{p}}) (\mathbf{p}_i- \Bar{\mathbf{p}})^{\top},
    \label{equ:cm}
\end{equation}
where $\Bar{\mathbf{p}} = \frac{1}{k}  \sum_{i=1}^{k} \mathbf{p}_i$, $k\geq 5$.
The eigenvectors $\mathbf{v_i}$ are estimated by $C \mathbf{v_i} = \lambda_i \mathbf{v_i}, i\in \{0,1,2\}$, where eigenvalues $\lambda_0, \lambda_1, \lambda_2$ are in the ascending order. The normal vector is aligned with the eigenvector corresponding to the smallest eigenvalue $\lambda_0$. For consistency in the direction of normal vectors, we apply $\mathbf{n}=-\mathbf{n}$ if $\mathbf{n}^{\top} (\mathbf{p}_i - \mathbf{p}_v) < 0$.

In selecting salient planar points, the surface curvature $c_{\mathbf{p}_i}$ of point $\mathbf{p}_i$ serves as a measure for evaluating the smoothness of the local surface. The surface curvature is defined as $c_{\mathbf{p}_i} = \frac{\lambda_0}{\lambda_0 + \lambda_1 + \lambda_2}$, where $\lambda_0 < \lambda_1 < \lambda_2$. We choose the planar feature point whose curvature is below the user-defined threshold $c_{th}$.

\subsubsection{Data Association}
The objective of our odometry scheme is to estimate the 6-DoF pose $\mathbf{T}_k \in \mathbb{R}^{4\times 4} \in \mathrm {SE}(3)$ of the current frame $k$ relative to the global mesh model. To facilitate the smoothness of the robot's motion, we use the constant velocity model to initialize the predicted pose $\widehat{\mathbf{T}}_{k}$. Given the previous estimated poses $\mathbf{T}_{k-2}$ and $\mathbf{T}_{k-1}$, the predicted pose of current frame is computed by $\widehat{\mathbf{T}}_{k} =  \mathbf{T}_{k-1}\mathbf{T}_{k-2}^{-1}\mathbf{T}_{k-1}$. 

For incoming point cloud frame $\mathcal{P}_k$, point clouds $  \mathbf{p}_{i} \in \mathcal{P}_k$ are transformed to the estimated global coordinate system  $ \mathbf{p}_{w} = \widehat{\mathbf{T}}_{k}\mathbf{p}_{i} \in \mathcal{P}_w$ based on the predicted pose $\widehat{\mathbf{T}}_{k}$. For every triangular facet $\mathcal{S}_i$, we estimate its normal vectors $\mathbf{n}_{\mathcal{S}_i}$ using the cross product from its vertices. 

During each iteration of the point-to-mesh ICP process, we establish correspondences between the point clouds $\mathbf{p}_w$ and triangular mesh $\mathcal{S}$ through nearest neighbor search. Moreover, $\mathbf{v}_{i}$ represents the point on the triangular facet $\mathcal{S}_i$. To ensure correct correspondences, two criteria are employed to filter outliers. Firstly, the point-to-mesh distance $|\mathbf{n}_{\mathcal{S}_i}^{\top} (\mathbf{p}_{w}-\mathbf{v}_{i}) |$ is utilized for determining the corresponding facet for the point $\mathbf{p}_{w}$. Besides, a threshold $c_s$ is set to constrain the absolute value of the cosine similarity $\frac{|\mathbf{n}_{\mathcal{S}_i}^{\top} \mathbf{n}_w |}{\|\mathbf{n}_{\mathcal{S}_i}\|\cdot\|\mathbf{n}_w\|}$ between the normals of the point and facet. Given the uncertainty in the orientation of the normal vectors, the absolute value of the cosine similarity is utilized to evaluate the alignment between their normals. Correspondences are chosen based on this measure. We select those matches whose absolute value of the cosine similarity exceeds the predefined threshold $c_s$. 

\subsubsection{Pose Optimization}
To achieve more effective convergence during the optimization process, we focus on estimating the relative pose $\mathbf{T}_{icp,k}$ instead of directly computing the global pose $\mathbf{T}_{k}$. $\mathbf{T}_{icp,k}$ is the deviation between prediction frame $\mathcal{P}_w$ and global triangular meshes. Thus, we aim to minimize the point-to-mesh error in the following:
\begin{equation}
\mathbf{T}_{icp,k} = \mathop{\arg\min}\limits_{\mathbf{T}_{icp,k}}  \sum_{(\mathbf{p}_w, \mathbf{v}_{\mathcal{S}_i}) \in \mathcal{C}} \|\mathbf{n}^\top_{\mathcal{S}_i}(\mathbf{T}_{icp,k}  \mathbf{p}_w - \mathbf{v}_{\mathcal{S}_i})\|_2  , 
\label{equ:optimization}
\end{equation}
where $\mathcal{C}$ is the set of appropriate correspondence between point and meshes, $\mathbf{n}_{\mathcal{S}_i}$ is the triangular facet normal,  and $\mathbf{v}_{\mathcal{S}_i}$ is a point cloud on the triangular facet $\mathcal{S}_i$. Our triangular mesh indicates the set of zeros of $I^{\mathcal{P}_k}$ by linear interpolation. If each point-to-mesh residual is taken into account separately, any point on the triangular facet can be chosen as $\mathbf{v}_{\mathcal{S}_i}$. Arbitrary selection of point $\mathbf{v}_{\mathcal{S}_i}$ may result in slow convergence of the optimization objective $\mathbf{T}_{icp,k}$.
To ensure the uniform convergence of the optimization process, we choose the point cloud on the plane as the $\mathbf{v}_{\mathcal{S}_i}$.

Equ.~\ref{equ:optimization} presents a nonlinear least squares minimization problem, which is typically solved through an iterative Gauss-Newton algorithm. The error term is defined as $e=\mathbf{n}^\top_{\mathcal{S}_i}(\mathbf{T}_{icp,k}  \mathbf{p}_w - \mathbf{v}_{\mathcal{S}_i})$, the Jacobian $\mathbf{J}$ is computed as follows:
\begin{equation}
\mathbf{J} = \frac{\partial e}{\partial \mathbf{T} }=
\mathbf{n}_{\mathcal{S}_i}^\top \begin{bmatrix}
  \mathbf{I} & -\left [{\mathbf{T}_{icp,k}\mathbf{p}_{w}} \right ]_{\times }
\end{bmatrix},
\end{equation}
where $\left [ \cdot \right ]_{\times }$ represents the skew symmetric matrix of a vector. For each iteration $j$, the pose increment $\Delta \mathbf{T}_{icp,k}^{j} \in \mathbb{R}^{6}$ is given by $\Delta \mathbf{T}_{icp,k}^{j} = (\mathbf{J}^{\top}\mathbf{J})^{-1}\mathbf{J}^{\top}e$, with $\mathbf{T}_{icp,k}^{j-1}$ representing the pose $\mathbf{T}_{icp,k}$ from iteration $j-1$. The relative pose $\mathbf{T}_{icp,k}^{j}$ is then updated as $\mathbf{T}_{icp,k}^{j} = \mathrm{Exp}(\Delta \mathbf{T}_{icp,k}^{j})\mathbf{T}_{icp,k}^{j-1}$, where $\mathrm{Exp}(\cdot)$ is mapping function from $\mathbb{R}^{6}$ to $\mathrm{SE}(3)$~\cite{sola2018micro}. When the increment is below a predefined threshold, the optimization converges and the final relative pose is $\mathbf{T}_{icp,k} = \mathbf{T}_{icp,k}^{j}$, where $j$ means the final iteration. Then the global pose is obtained as $\mathbf{T}_{k} = \mathbf{T}_{icp,k} \widehat{\mathbf{T}}_{k}$, which transforms the current scan into the global coordinate system for incremental voxel meshing.

\subsection{Incremental Voxel Meshing}\label{Incremental Voxel Meshing}
To achieve real-time mapping for large-scale environments, we propose a two-stage incremental voxel meshing method. Firstly, an efficient hybrid-weighted voxel fusion is proposed, which uses sparse voxels to preserve global map information and allows traversing each point only once to update every scan. Secondly, we leverage height-adaptive voxel blocks to compress space and extract surface mesh efficiently. 

\subsubsection{Implicit Function}
We take the Implicit Moving Least Squares (IMLS), an excellent implicit function $f:\mathbb{R}^3 \rightarrow \mathbb{R}$,  to approximate the underlying surface. 
The reconstruction surface is implicitly defined to the zero level-set $Z(f)$. 
For each voxel $\mathbf{v} = (x,y,z)^{\top}$, the IMLS~\cite{kolluri2008provably} function is defined as below
\begin{equation}
     I^{\mathcal{P}_k}(\mathbf{v}) = \frac{\sum_{\mathbf{p}_i \in \mathcal{P}_k} w_{i}^H (\mathbf{v})(\mathbf{n}_{i}^{\top} (\mathbf{v}-\mathbf{p}_{i})) }
    {\sum_{\mathbf{p}_j \in \mathcal{P}_k}w_{j}^H(\mathbf{v})} ,
    \label{equ:imls1}
\end{equation}
where $\mathcal{P}_k$ is the point cloud map of the current frame $k$, $\mathbf{p}_{i}  \in \mathcal{P}_k$ is a point, $w_{i}^H(\mathbf{v})$ is hybrid weight, and $\mathbf{n}_{i} = (n_x,n_y,n_z)^{\top}$ is the normal of point $\mathbf{p}_{i}$.

Despite its advantages in scalability to large datasets, and robustness to input noise~\cite{kolluri2008provably}, applying IMLS in large-scale scenes usually results in significant time consumption. To speed up the IMLS calculation, an improved solution is to construct a k-d tree for $\mathcal{P}_k$. Unfortunately, frequently building a k-d tree is computationally intensive with the time complexity of $O(N\log N)$ per tree construction, where $N$ is the number of point clouds in $\mathcal{P}_k$. Our voxel fusion improves the IMLS calculation without extra data structure for $\mathcal{P}_k$.

\subsubsection{Voxel Representation}
We make use of sparse voxels to retain the information of accumulative point clouds, including a position, normal, SDF value, weight, and frame index. 
The voxel is defined as the minimal unit with the same size in space. Given the volumetric resolution $(r_x, r_y, r_z)$, a point $\mathbf{p} = (p_x,p_y,p_z)^{\top} \in \mathcal{P}_w$, point cloud map in the global coordinate system, corresponds to a voxel as below
\begin{equation}
    \begin{pmatrix} v_x \\ v_y \\ v_z \end{pmatrix}
    = 
    \begin{pmatrix} \lfloor p_x/r_x \rfloor \\ \lfloor p_y/r_y \rfloor \\ \lfloor p_z/r_z  \rfloor \end{pmatrix} . 
    \label{equ:voxel}
\end{equation}

To distinguish our computational model for SDF estimation from others, we refer to our voxel as ``passive voxel".
For simplicity, unless specified, we will use the term ``voxel" as ``passive voxel" in the following.
   \begin{figure}[htbp]
      \centering
    \begin{minipage}{0.25\linewidth}
		 \centerline{\includegraphics[width=\linewidth]{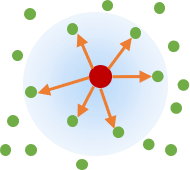}}
        \centerline{(a) Active way}
	\end{minipage}
 \hspace{0.12\linewidth}
	\begin{minipage}{0.25\linewidth}
         \centerline{\includegraphics[width=\linewidth]{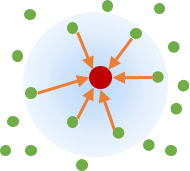}}
        \centerline{(b) Passive way}
	\end{minipage}
      \caption{Computational models for SDF estimation in (a) Active way and (b) Passive way. The green points denote the point clouds. The red point is the voxel. The blue circle represents the neighborhood centered around the voxel.
      }
      \label{fig:voxel}
   \end{figure}

Different from actively searching for points around itself~\cite{deschaud2018imls}, our proposed scheme emphasizes the point-to-voxel increments.
Fig.~\ref{fig:voxel} illustrates the computational models used to estimate the signed distance function (SDF) values in active and passive ways. Specifically,
the voxels $\{\mathbf{v}_0^A,..., \mathbf{v}_M^A\}$ in active way find the neighboring points to calculate the implicit function, which leads to substantial computation in searching among enormous point clouds $\{\mathbf{p}_0,...,\mathbf{p}_N\}$ with the time complexity of $O(NM)$.
Moreover, as the scene expands, the number of voxels $M$ increases.
In contrast to actively finding the neighbors, our passive voxels $\{\mathbf{v}_0^P,..., \mathbf{v}_M^P\}$ receive the SDF increments $ \mathbf{n}_{i}^{\top} (\mathbf{v}-\mathbf{p}_{i})$ from the surrounding points, which avoids the extensive searches. A point can determine the voxel influenced by itself via Equ.~\ref{equ:voxel}. In other words, the passive voxel representation can be built by searching the nearest points at the time complexity of $O(1)$. Consequently, the overall time complexity of our proposed scheme is $O(N)$, where $N$ is the total number of scattered points.
Our proposed method reduces the time complexity from $O(NM)$ to $O(N)$ and improves efficiency without requiring any extra efficient representation of the point clouds.

\subsubsection{Hybrid-weighted Voxel Integration}
Our voxel emphasizes the incremental fusion from point to itself. 
The point-to-voxel increment is computed as follows:
\begin{equation}
     d_{i}(\mathbf{v}) = \mathbf{n}_{i}^{\top} (\mathbf{v}-\mathbf{p}_{i}) ,
    \label{equ:paI2}
\end{equation}

\noindent
where $d_{i}(\mathbf{v})$ is the distance between the voxel and local surface formed by point $\mathbf{p}_i$.

We make use of a hybrid weight that describes the distance and normal feature similarity. For point $\mathbf{p}_{i}$, the hybrid weight $w_{i}^H (\mathbf{v})$ is defined as follows:

\begin{equation}
     w_{i}^H (\mathbf{v}) =   e^{ - \frac{||\mathbf{v}-\mathbf{p}_{i}||^2}{h} } + \lambda_n \mathbf{n}_{i}^{\top}\mathbf{n}_{\mathbf{v}}  ,
    \label{equ:imls2}
\end{equation}

\noindent
where $h$ is the range parameter. $\lambda_n$ is the weight. $\mathbf{n}_{i}$ is the normal of point $\mathbf{p}_{i}$, and $\mathbf{n}_{\mathbf{v}}$ is the normal of voxel $\mathbf{v}$.

Our real-time LiDAR mapping can be considered as fitting surfaces from a continuous stream of point clouds. For point clouds $\{\mathbf{p}_0,...,\mathbf{p}_n\}$ related to the voxel $\mathbf{v}$, SDF value in the voxel is updated as in~\cite{curless1996volumetric}:
\begin{equation}
     I(\mathbf{v}) \leftarrow I(\mathbf{v}) + \frac{w_{n}^H(\mathbf{v})(d_{n}(\mathbf{v})-I(\mathbf{v}))}{W(\mathbf{v})+w_{n}^H(\mathbf{v})},
    \label{equ:Iupdate1}
\end{equation}
\begin{equation}
     W(\mathbf{v}) \leftarrow W(\mathbf{v})+w_{n}^H(\mathbf{v}),
    \label{equ:Iupdate1}
\end{equation}

\noindent where $I(\mathbf{v})$ and $W(\mathbf{v})$ are the SDF value and weight of the voxel, respectively. $d_{n}(\mathbf{v})$ is the distance between the voxel and local surface formed by point $\mathbf{p}_n$, and 
$w_{n}^H(\mathbf{v})$ is its weight.

    \begin{figure}[thpb]
    \centering
        \centerline{\includegraphics[width=0.25\textwidth]{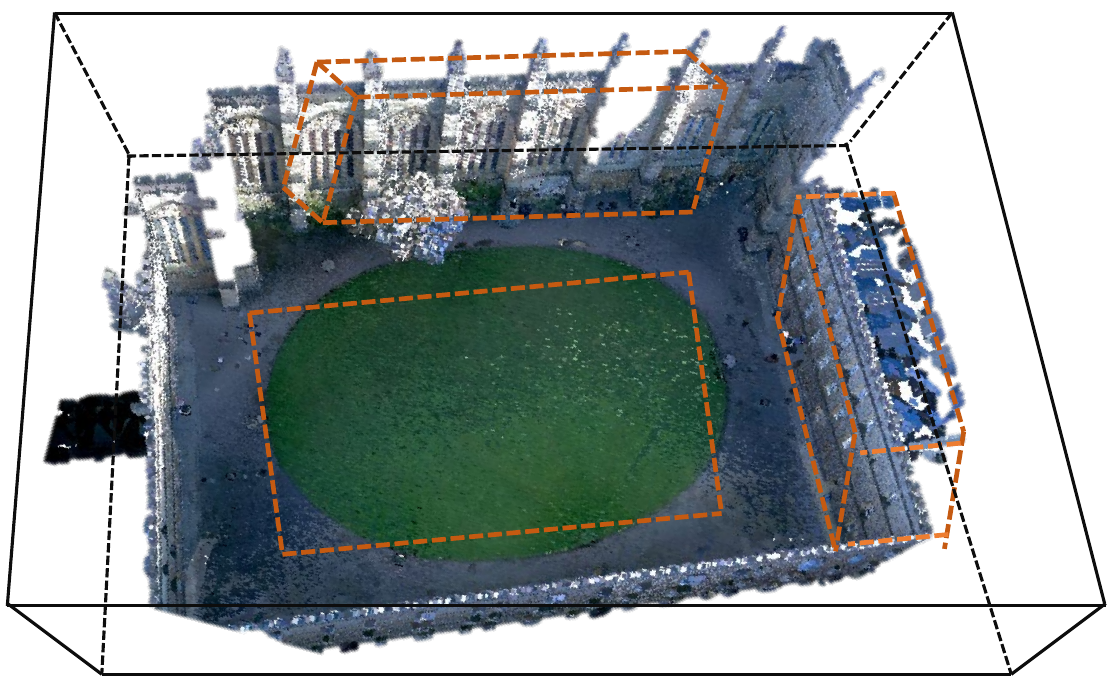}} 
    	\caption{The black cuboid is a predetermined bounding box. The orange cuboid is our voxel block. 
     }
    	\label{fig:block}
    
   \end{figure}

Due to the sparsity of the point clouds, each point individually affecting its corresponding voxel is insufficient. The influence of the point on neighboring voxels should also be considered. To this end, we set an influence region centered on the selected voxel as a square with the edge length of $l$, where $l$ is a small constant. 

Note that hybrid-weighted voxel integration focuses on point-to-voxel increments, which is suitable for parallel acceleration. Moreover, the fusion completes the update by traversing all points only once and allows for the safe removal of data from previous frames. This not only significantly improves efficiency but also reduces memory consumption.

\subsubsection{Partitioned Meshing}
Our sparse voxels are dispersed throughout space. To acquire the voxel information within a specific space, a complete traversal of the range is required. Our approach allows quick access at each position including voxel and 
empty space. However, it is inefficient and unnecessary to check a large number of empty locations. To reduce the empty space, we divide the 3D volume into voxel blocks with the same length $L_{vb}$, same width $W_{vb}$, and variable height $H_{vb}$. Conceptually, the voxel block contains a certain amount of voxels, which is only located around the reconstructed surfaces, as shown in Fig.~\ref{fig:block}. The voxel block $\mathcal{B}$ is a scalable cuboid that dynamically adjusts its height to match the local scene in 3D space. The height is computed as follows

\begin{equation}
     H_{vb}= \mathop{\max}_{\mathcal{B}} z -  \mathop{\min}_{\mathcal{B}} z .
    \label{equ:voxelblock}
\end{equation}

\noindent
Height-adaptive voxel blocks are designed to compress space and eliminate the requirement for a predefined bounding box.

The block-based Signed Distance Function (SDF) map provides quick access, allowing for parallel extraction of the triangular mesh using the marching cubes algorithm~\cite{lorensen1987marching}.
As the neighboring frames are similar, performing repetitive surface extraction on each scan may require a substantial computational cost. To this end, we extract the explicit surface at intervals $t_s$. We suggest that $t_s$ is constrained by the total number of voxel blocks. Therefore $t_s$ becomes a dynamic parameter that does not require manual setting. 
By exploiting the benefits of our proposed hybrid-weighted voxels fusion method and our easily accessible blocked-based SDF map, we can incrementally reconstruct the surface meshes.

\subsection{Parallel Spatial-hashing Scheme}\label{voxelhashing}
To achieve the parallelization of voxel operations, we employ a simple and efficient spatial hash-based scheme. Besides, our proposed voxel deletion scheme enables long-term reconstruction and ensures the mesh quality involved with little influence.

\subsubsection{Parallel Hash Table}
The world coordinates of voxels are mapped to their corresponding hash codes using a spatial hash function. As in~\cite{teschner2003hash}, a spatial hashing function $f:\mathbb{R}^3 \rightarrow \mathbb{R}$ is defined for a voxel position $\mathbf{v} = (x,y,z)$:
\begin{equation}
     H(x,y,z) = (x \cdot p_1 \oplus y \cdot p_2 \oplus z \cdot p_3) \bmod n
 ,
    \label{equ:hashing}
\end{equation}
\noindent
$p_1$, $p_2$, $p_3$ are large prime numbers, which are set to 73856093, 19349663, 83492791, respectively. The value $n$ is the size of the hash table.

Although hash-based voxel representation is a common solution for processing and analyzing 3D scenes due to its compactness and flexibility, it is usually implemented on CPUs.
To maximize its efficiency, our spatial-hashing scheme is implemented on GPU. Unfortunately, our hybrid voxel fusion leads to data races~\cite{artho2003high} on GPU. To this end, distributed locks are employed to ensure those threads with serial operations are implemented mutually exclusive writing to memory. This achieves parallel operations among voxels while also supporting serial operations within them.
   
\subsubsection{Collision Resolving Strategy}
Collisions in the hash table may lead to many invalid values of the reconstructed mesh, thus not fully utilizing the storage space. To deal with hash collisions, we employ open addressing like robin hood hashing~\cite{celis1985robin}, linear probing, and quadratic probing. In our experiments, linear probing is chosen as the collision resolving strategy empirically. 

\subsubsection{Voxel Deletion Scheme}
A key observation is that the initial scene information typically remains unchanged with the extension of the scene. In this perspective, we propose the voxel deletion scheme, which converts the invariant information stored in the hash table into explicit mesh surfaces and deletes the voxels that provide the overdue information. Due to the voxel removal, the consecutive new voxels can be safely inserted. Our proposed voxel deletion scheme enables the continuous reconstruction of the large-scale surface mesh within the limited memory.

\subsubsection{Voxel Operations}
The position of a voxel serves as the unique identification. To insert a new hash item, we calculate its hash code by Equ.~\ref{equ:hashing}. When the corresponding entry in the hash table is empty, we directly insert the voxel into it. If the entry has been occupied, we employ a collision resolving strategy to obtain a new mapping value. Then, we iterate over the above steps until the voxel is inserted. Algorithm~\ref{algo1} describes the parallel insertion of voxels through atomic operations on mutexes.

Since the sequential LiDAR scans are processed, voxel information needs to be frequently updated.
To retrieve and update a voxel, we calculate its hash code through Equ.~\ref{equ:hashing} and locate the corresponding entry in the hash table. 
Then, we compare the unique identification, voxel position, of two voxels. 
If they are the same, we find the targeted voxel and update it as described in Section~\ref{Incremental Voxel Meshing}.
Otherwise, we obtain a new mapping value by a collision resolving method and literately repeat the above steps until the position matches. 

 \begin{algorithm}[htbp]
    \small
    \caption{Voxel Insertion Algorithm}
    \label{algo1}
    \begin{algorithmic}

    \STATE $//$ for each point $ \mathbf{p} = (x,y,z)^{\top}$ simultaneously do

    \STATE blocked $\leftarrow$ false ;
    \STATE $s = Hash(V(x,y,z))$ ;

    \FOR{$i \leftarrow 1$ to $T_{max}$}
    \STATE blocked $\leftarrow$ true;
        \WHILE {blocked is true}
            \IF{Mutex(s) = unlock and Mutex(s) $\leftarrow$ lock}
                \IF{voxel $\mathbf{v}_s$ is empty}
                    \STATE insert the voxel;  
                \ELSIF{KEY$(\mathbf{v}_s)$ = $V(x,y,z)$}
                    \STATE exit;
                \ENDIF
                \STATE Mutex(s) $\leftarrow$ unlock;
                \STATE blocked $\leftarrow$ false;
             \ENDIF
        \ENDWHILE 
        \STATE $s \leftarrow s + f(i)$;
    \ENDFOR
    \end{algorithmic}
    \end{algorithm}
    
To delete an existing voxel, we retrieve the position of its corresponding hash item at the first step. Then, we clean up this hash item. 

\subsubsection{Voxel Block Implementation}
For a voxel block $\mathbf{v}_b = (x_b,y_b,z_b)^{\top}$, we construct a hash function $f:\mathbb{R}^2 \rightarrow \mathbb{R}$ using the first two components of the vector. Specifically $ H(x,y) = (x \cdot p_1 \oplus y \cdot p_2) \bmod n$, as in Equ.~\ref{equ:hashing}. The robin hood hashing~\cite{celis1985robin} is used to resolve the collisions. 
Voxel block is able to effectively compress the space, which can be used in scenes with unknown volume. According to the voxel deletion scheme, we delete those expired voxels so that the expired voxel blocks can also be removed.

    \begin{figure*}[thpb]
    \centering
        \centerline{\includegraphics[width=0.95\textwidth]{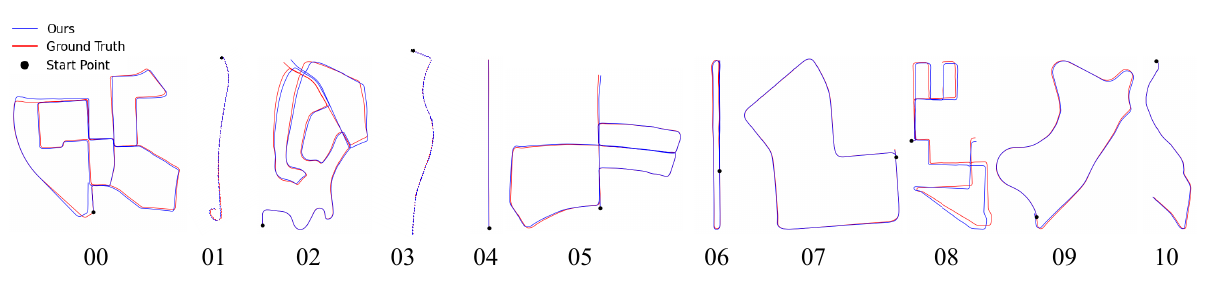}} 
    	\caption{Trajectories estimated by our Mesh-LOAM on the KITTI sequences `00' to `10'.  }
    	\label{fig:traj_kitti}
    
   \end{figure*}
   
\begin{table*}
\centering

\caption{\small Results on KITTI Odometry Benchmark}
\label{tab:kitti}
\resizebox{0.85\linewidth}{!}{
\begin{threeparttable}
\begin{tabular}{ccccccccccccccc}
\toprule

 \multicolumn{1}{l}{Map type} & \multicolumn{2}{c}{Method} & 00    & 01    & 02    & 03    & 04    & 05    & 06    & 07    & 08    & 09    & 10    & Mean \\
    \midrule
    
    \multirow{4}[4]{*}{Surfel} & \multicolumn{2}{c}{\multirow{2}{*}{Suma \cite{behley2018suma}}}& {0.70}& 1.70& 1.10& 0.70& {0.40}& 0.50& 0.40& 0.40& 1.00& {0.50}& 0.70& 0.74 \\
           &  &  & 0.30& 0.30& 0.40& 0.50& 0.30& 0.20& 0.20& 0.30& 0.40& 0.30& 0.30& 0.32 \\
\cmidrule{2-15}          & \multicolumn{2}{c}{\multirow{2}{*}{Suma++ \cite{chen2019suma++}}}& 0.64& 1.60& 1.00& 0.67& 0.37& 0.40& 0.46& 0.34& 1.10& 0.47& 0.66& 0.70 \\
           &  &  & 0.22& 0.46& 0.37& 0.46& 0.26& 0.20& 0.21& 0.19& 0.35& 0.23& 0.28& 0.29 \\
    \midrule
    \midrule
    \multirow{2}[2]{*}{NDT} & \multicolumn{2}{c}{\multirow{2}{*}{Litamin2 \cite{yokozuka2021litamin2}}}& 0.70& 2.10& 0.98& 0.96& 1.05& 0.45& 0.59& 0.44& 0.95& 0.69& 0.80& 0.88 \\
           &  &  & 0.28& 0.46& {0.32}& 0.48& 0.52& {0.25}& 0.34& 0.32& 0.29& 0.40& 0.47& 0.38 \\
    \midrule
    \midrule
    \multirow{6}[4]{*}{Point Cloud} & \multicolumn{2}{c}{\multirow{2}{*}{FLOAM~\cite{wang2021f}}} &
0.73 & 0.95 & 0.83 & 1.47 & 0.50 & 0.63 & 0.48 & 0.71 & 1.03 & 0.66 & 0.94 & 0.81 \\
 & & & 0.44 & 0.16 & 0.34 & 0.35 & 0.26 & 0.42 & 0.31 & 0.66 & 0.40 & 0.30 & 0.42 & 0.37 \\
\cmidrule{2-15}          & \multicolumn{2}{c}{\multirow{2}{*}{KISS-ICP~\cite{vizzo2023kiss}}} &
 \textbf{0.51} & 0.72 & 0.53 & 0.66          & 0.35 & \textbf{0.30} & \textbf{0.26} & 0.33 & \textbf{0.81} & 0.49 & 0.56 & \textbf{0.50} \\
& & & \textbf{0.19} & 0.11 & 0.15 & \textbf{0.16} & 0.14 & \textbf{0.14} & 0.08          & 0.16 & \textbf{0.18} & 0.13 & 0.18          & 0.15          \\
\cmidrule{2-15}          & \multicolumn{2}{c}{\multirow{2}{*}{Ours (point-to-plane)\tnote{1}}} &
 0.64 & 2.32 & 0.58 & 0.63 & \textbf{0.31} & 0.39 & 0.36 & 0.35 & 0.91 & 0.53 & 0.78 & 0.71 \\
& & & 0.23 & 0.17 & 0.19 & 0.25 & 0.11 & 0.18 & 0.08 & 0.25 & 0.28 & 0.16 & 0.24 & 0.19 \\
    \midrule
    \midrule
    \multirow{6}[4]{*}{Mesh} & \multicolumn{2}{c}{\multirow{2}{*}{Puma \cite{vizzo2021puma}}}& 1.46& 3.38& 1.86& 1.60& 1.63& 1.20& 0.88& 0.72& 1.44& 1.51& 1.38& 1.55 \\
           &  &  & 0.68& 1.00& 0.72& 1.10& 0.92& 0.61& 0.42& 0.55& 0.61& 0.66& 0.84& 0.74 \\          
\cmidrule{2-15}       &   \multicolumn{2}{c}{\multirow{2}{*}{SLAMesh~\cite{slamesh}}} & 0.77 & 1.25  & 0.77  & 0.64 & 0.50  & 0.52   & 0.53 & 0.36  & 0.87  & 0.57 & 0.65 & 0.68  \\
 &  &  & 0.35  & 0.30  & 0.30  & 0.43 & 0.13  & 0.30  & 0.22  & 0.23  & 0.27 & 0.25 & 0.42  & 0.29   \\
\cmidrule{2-15}       &   \multicolumn{2}{c}{\multirow{2}{*}{Ours}} & 0.53 & \textbf{0.64} & \textbf{0.52} & \textbf{0.63} & 0.41          & 0.34 & 0.35          & \textbf{0.31} & 0.82 & \textbf{0.47} & \textbf{0.56}          & 0.51          \\
 &  &  & 0.20 & \textbf{0.11} & \textbf{0.15} & 0.20          & \textbf{0.07} & 0.15 & \textbf{0.07} & \textbf{0.15} & 0.19 & \textbf{0.13} & \textbf{0.16} & \textbf{0.14} \\
\bottomrule
\end{tabular}

\begin{tablenotes}
\item[1] The point-to-plane loss is used to optimize the estimated poses.
\end{tablenotes}

\end{threeparttable}
}
\vspace{-0.2in}
\end{table*}

         \begin{figure}[thpb]
      \centering
      \includegraphics[width=\linewidth]{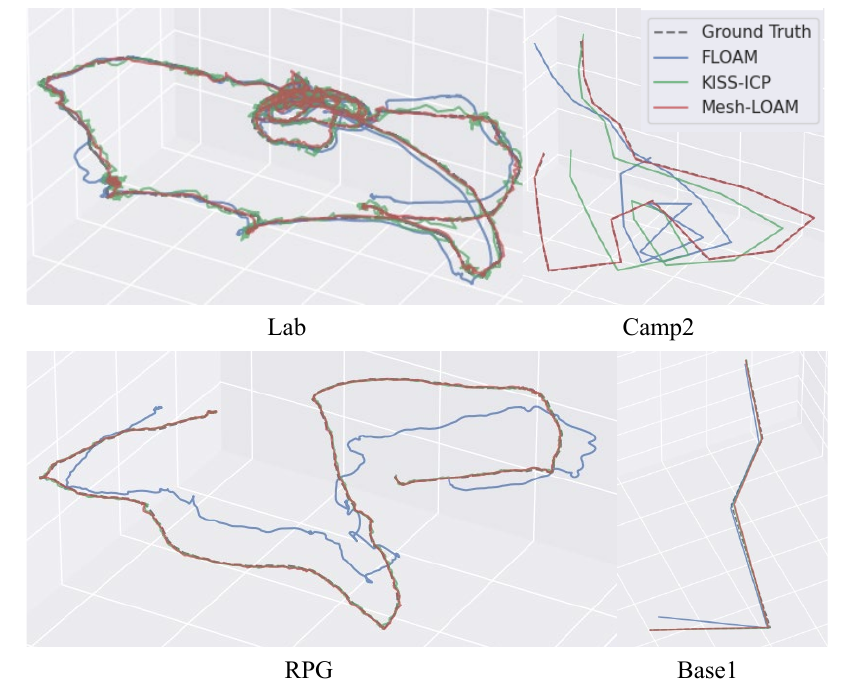}
      \caption{Comparisons on the Hilti SLAM challenge dataset
      }
      \label{fig:traj_hilti}
   \end{figure}

\section{EXPERIMENTS}  
In this section, we present the details of our experiments and evaluate our odometry and mapping approach on four real-world large-scale public datasets. Moreover, we show promising quantitative and qualitative results compared to the state-of-the-art approaches.
Additionally, we examine the effectiveness of our proposed point-to-mesh odometry as well as the voxel deletion scheme and discuss computational time.

\subsection{Implementation Details}
Our method is implemented in C++ with CUDA and primarily utilizes GPU, except for data transmission.
All experiments are conducted on a PC with an Intel Core i7-9800X CPU @ 3.80GHz and an NVIDIA GeForce RTX 2080Ti graphics card with 11GB GPU RAM.
We select the normals by the curvature threshold $c_{th}=0.1$. We set the voxel size as $r_x=r_y=r_z=0.1$.
When involved with the influence region for voxels, the edge length is set to $l=3$, meaning three times the size of the voxel.
We compute the hybrid-weighted IMLS using the parameters $h=0.05$ and $\lambda_n=0.2$. 
We set the maximum iteration $T_{max}=10$ in the voxel operations. For every voxel block, we set the length $L_{vb}=2$ and width $W_{vb}=2$. When choosing correct correspondences to compute point-to-mesh residuals, the threshold $c_s$ is set to 0.98.

\subsection{Datasets}
\textbf{KITTI}~\cite{geiger2012kitii}: The KITTI odometry~\cite{geiger2012kitii} is a large-scale outdoor dataset collected by Velodyne HDL-64E S2 LiDAR, which contains various street environments. Our experiments were conducted with Sequence `00' to `10', which have a total number of 23201 scans.
We made use of the KITTI odometry dataset to evaluate the odometry accuracy and map quality of our proposed approach. To examine the effectiveness of our method,
we also carried out an ablation study of point-to-mesh odometry and the voxel deletion scheme.

\textbf{HILTI 2021}~\cite{helmberger2022hilti}: To evaluate our proposed point-to-mesh odometry, we utilized the newer Hilti 2021 dataset. This dataset contains a series of real-world indoor sequences, including offices, labs, and construction environments, as well as outdoor sequences from construction sites and parking areas.
The handheld sensor platform used to record the data provides millimeter-accurate ground truth data for each sequence and includes an Ouster OS0-64 LiDAR, which collects long-range point cloud data with a 360$^{\circ}$ Fov and at a rate of 10 Hz.

\textbf{Mai City}~\cite{vizzo2021puma}: To qualitatively and quantitatively evaluate the recovered mesh quality of our proposed method, our experiments were conducted in the Mai City, which is a synthetic dataset generated from an urban-like scenario with a virtual Velodyne HDL-64 LiDAR. 

 \begin{figure*}[thpb]
      \centering 
      \begin{minipage}{0.24\linewidth}
        \centerline{\includegraphics[width=\textwidth]{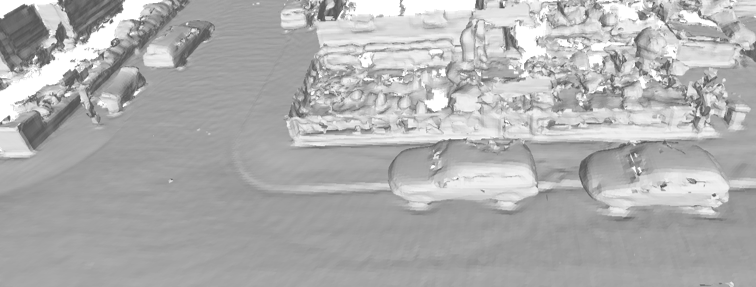}}
        \centerline{0.1}
	\end{minipage}
	\begin{minipage}{0.24\linewidth} \centerline{\includegraphics[width=\textwidth]{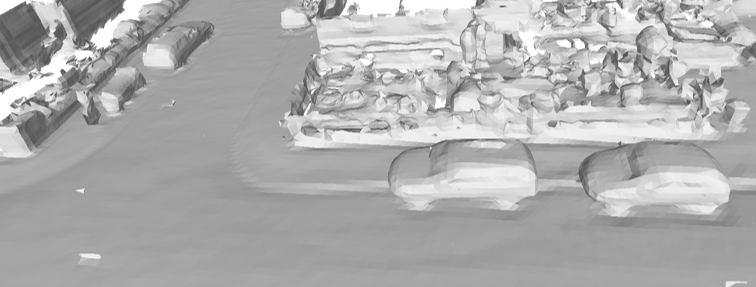}}
        \centerline{0.2}
	\end{minipage}
	\begin{minipage}{0.24\linewidth}
		 \centerline{\includegraphics[width=\textwidth]{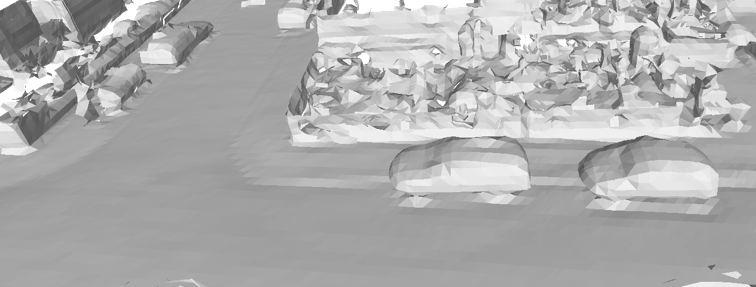}}
          \centerline{0.3}
	\end{minipage}
        \begin{minipage}{0.24\linewidth}
		 \centerline{\includegraphics[width=\textwidth]{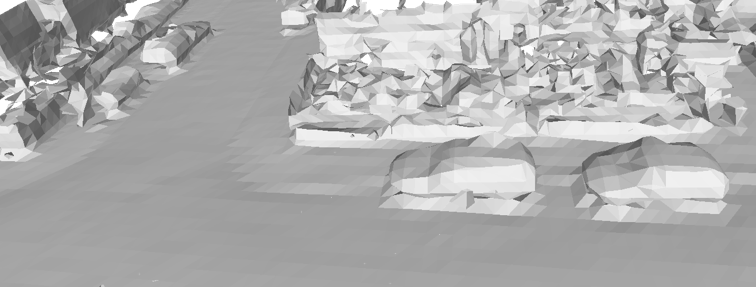}}
          \centerline{0.4}
	\end{minipage}
	\caption{Mapping results on Sequence `07' of the KITTI dataset with various resolutions from 0.1m to 0.4m. }
    	\label{fig:kitti07}
   \end{figure*}

\begin{figure*}[thpb]
      \centering 
      \begin{minipage}{0.24\linewidth}
		 \centerline{\includegraphics[width=\textwidth]{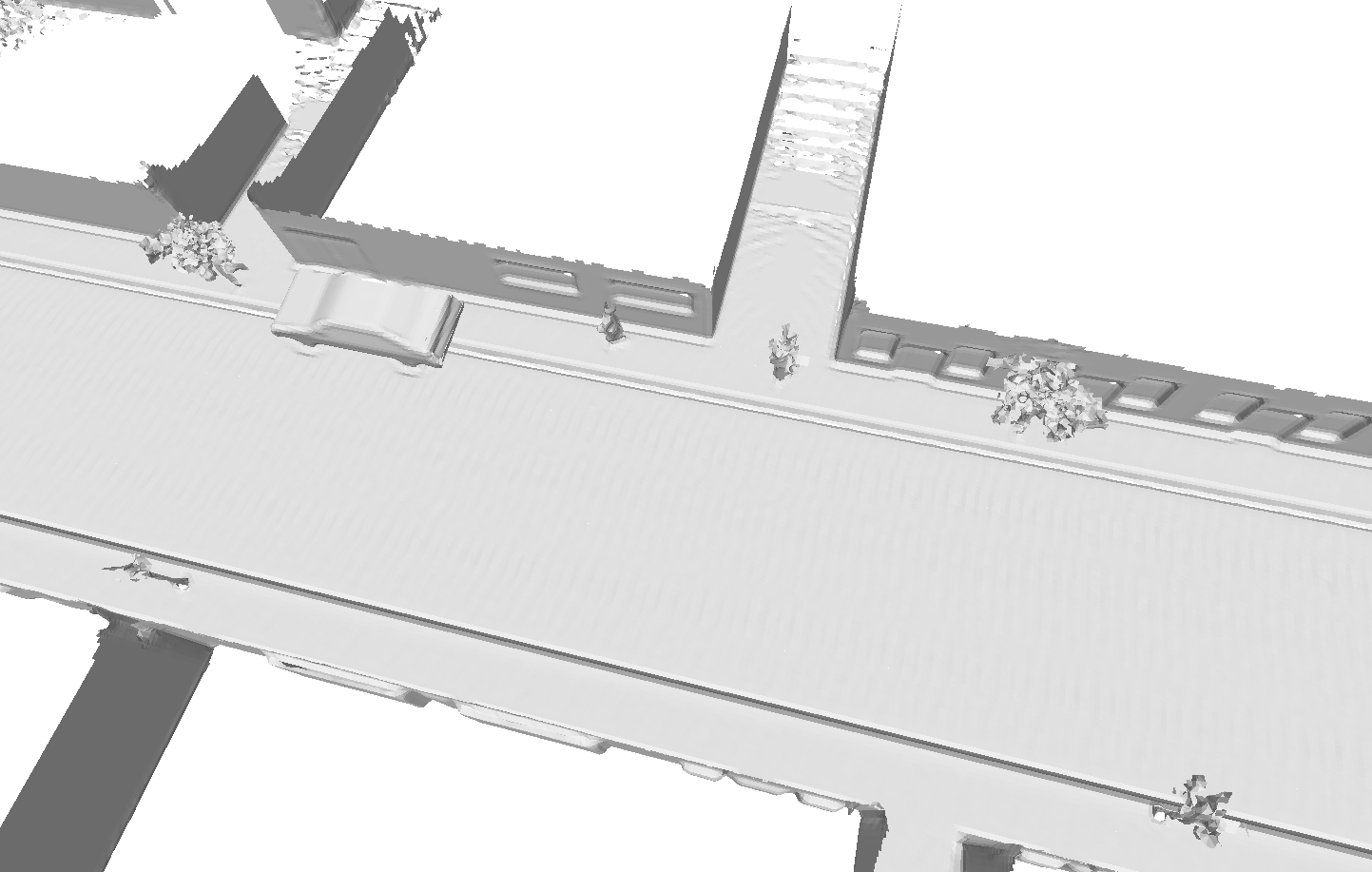}}
		\centerline{0.1}
	\end{minipage}
	\begin{minipage}{0.24\linewidth}
         \centerline{\includegraphics[width=\textwidth]{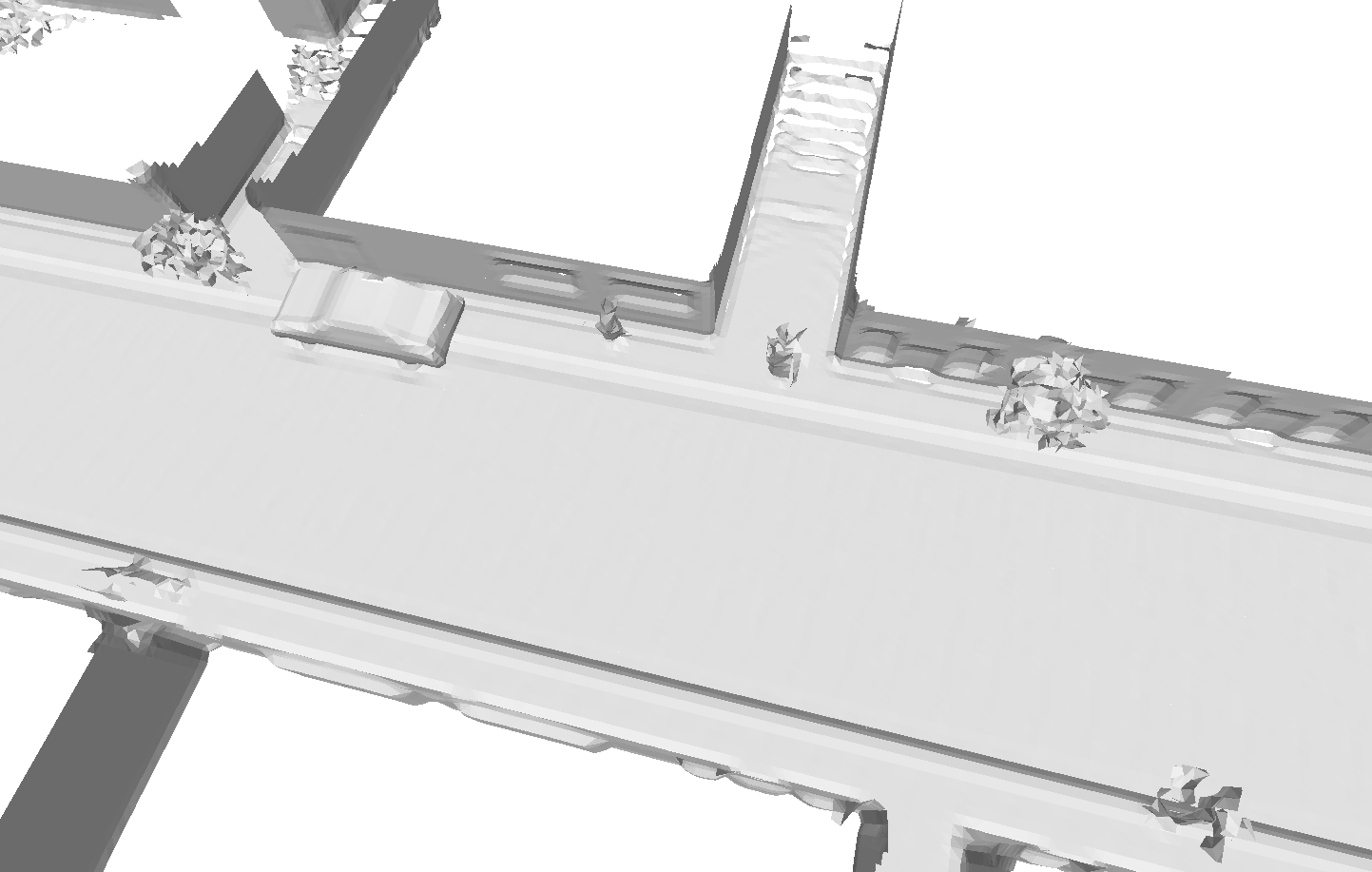}}
		\centerline{0.2}
	\end{minipage}
	\begin{minipage}{0.24\linewidth}
		 \centerline{\includegraphics[width=\textwidth]{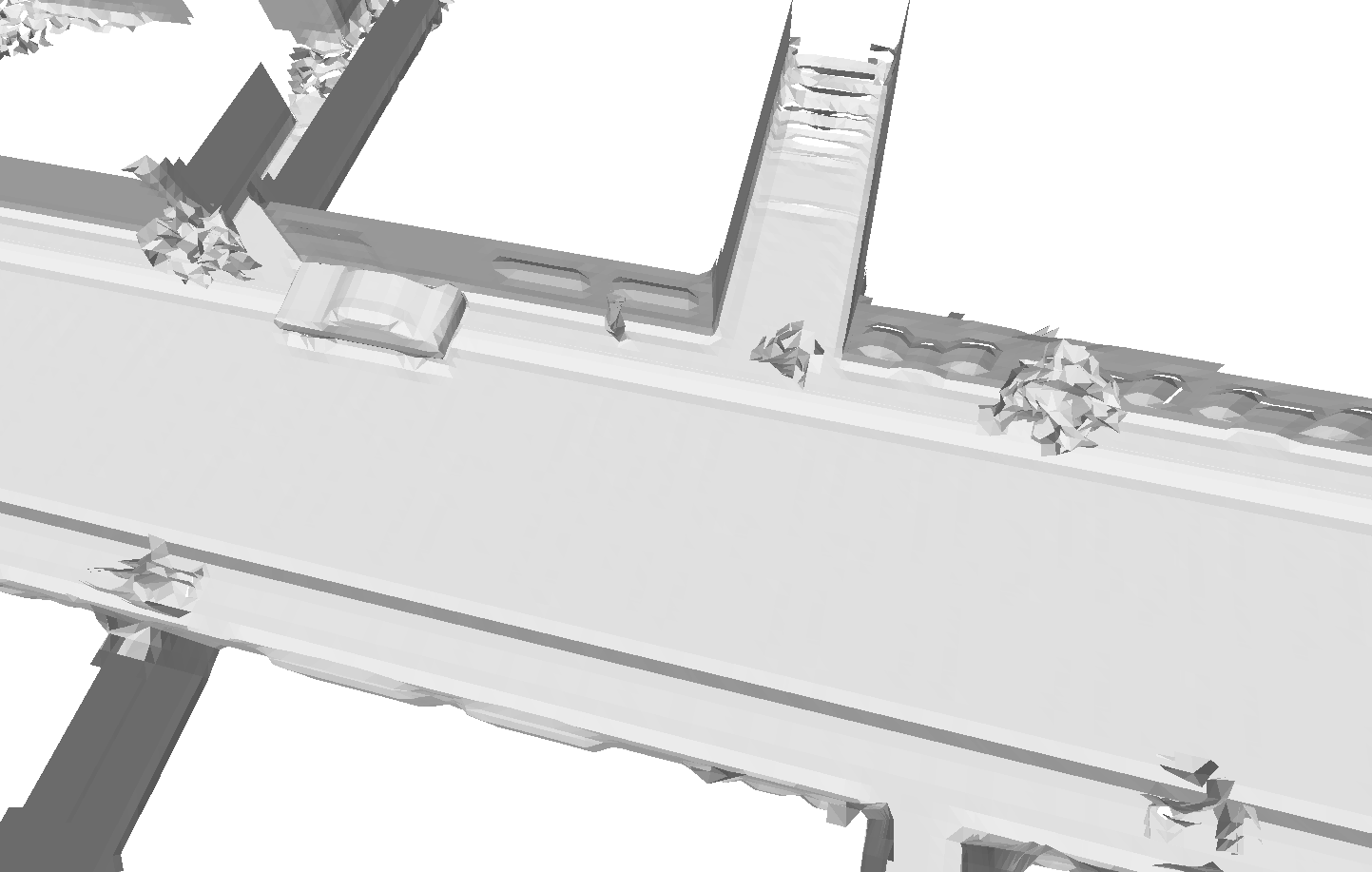}}
		\centerline{0.3}
	\end{minipage}
        \begin{minipage}{0.24\linewidth}
		 \centerline{\includegraphics[width=\textwidth]{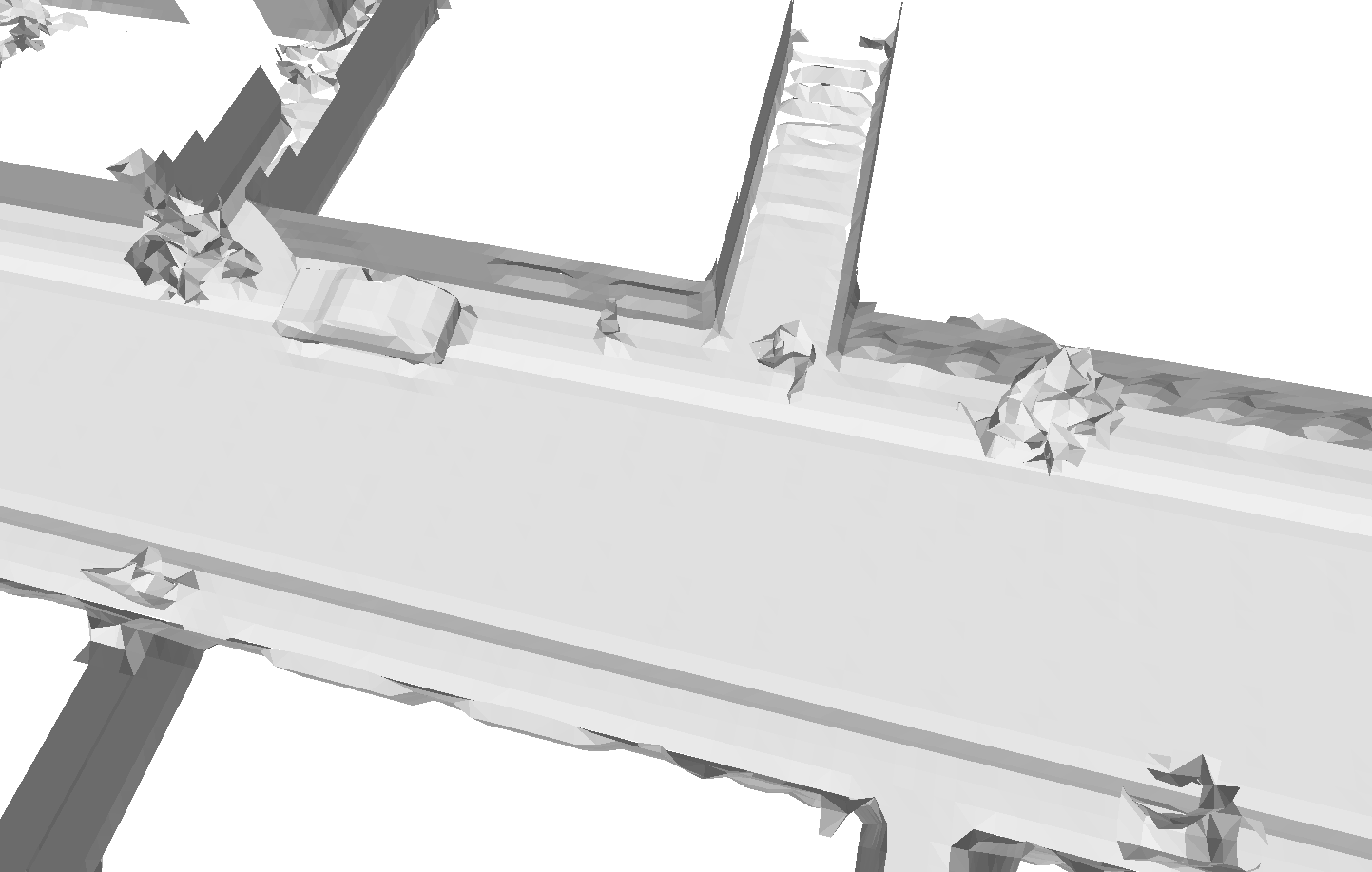}}
		\centerline{0.4}
	\end{minipage}
	\caption{Mapping results on the MaiCity dataset with various resolutions from 0.1m to 0.4m. }
	\label{fig:maicity}
   \end{figure*}


\begin{table}[t]
\centering
\caption{{\small ATE (m) on the Hilti SLAM Challenge Dataset}}
\resizebox{0.5\textwidth}{!}{
\begin{threeparttable}
\begin{tabular}{@{}lccccccc@{}}
\toprule
 Approach& Sensor& RPG 
  & Base1
 & Base4 & Lab
 & Cons2 & Camp2  \\
\midrule
 FLOAM~\cite{wang2021f} & \multirow{3}{*}{OSO-64} & 2.775&0.914&0.287&0.182&11.515& 8.946\\
 KISS-ICP~\cite{vizzo2023kiss} & &0.187&0.294&\bf{0.119}&0.073&0.835&5.052\\
 Ours && \bf{0.173} &  \bf{0.165}& 0.267& \bf{0.049}&\bf{0.083}&\bf{0.113}\\
\bottomrule
\end{tabular}
\end{threeparttable}
}
\label{tab:hilit}
\vspace{-0.2in}
\end{table}

       \begin{figure}[thpb]
      \centering
      \includegraphics[width=\linewidth]{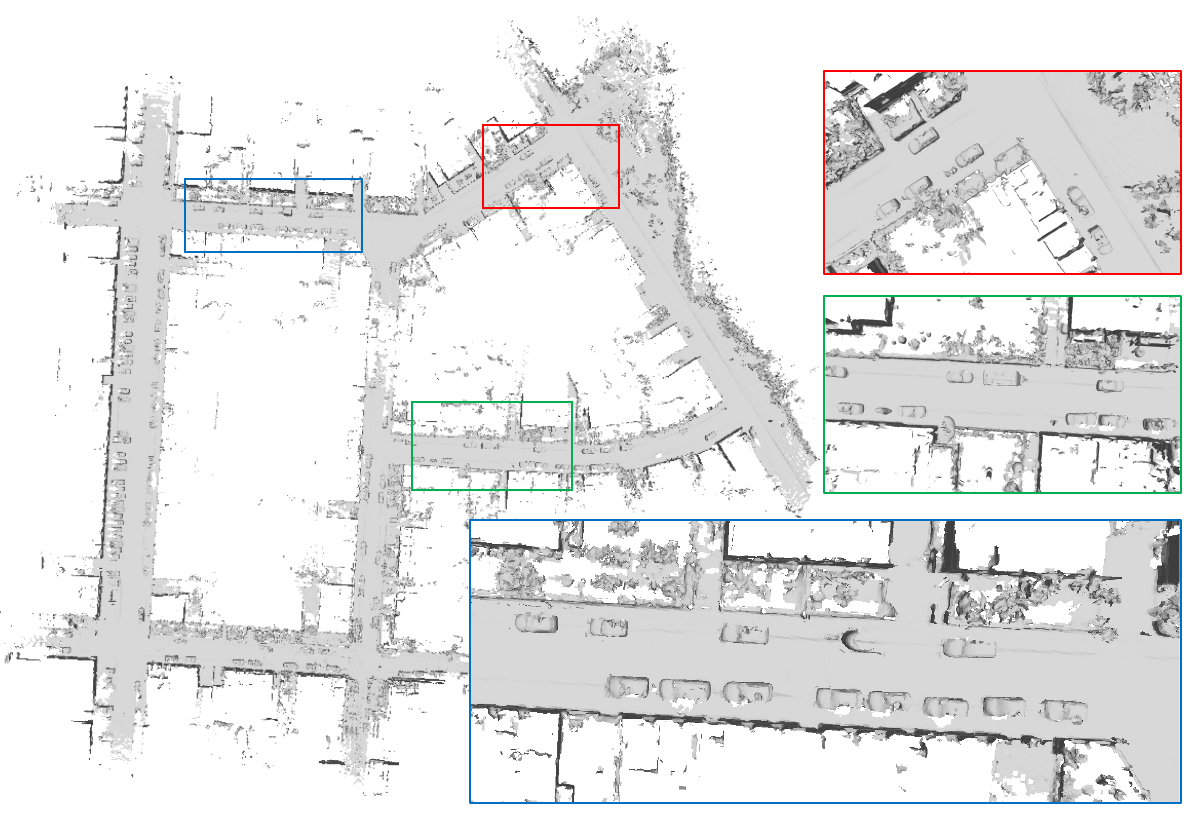}
      \caption{The mesh result of our proposed Mesh-LOAM approach with local details on KITTI sequence `07'.
      }
      \label{fig:reconstruction_kitti07}
   \end{figure}

\textbf{Newer College}~\cite{ramezani2020newer}: The Newer College dataset is collected by a handheld device with a multi-beam 3D LiDAR at Oxford University having a detailed millimeter-accurate 3D map. We quantitatively evaluate the recovered mesh quality on the two latter datasets with accessible ground truth mesh information.

\subsection{Odometry Evaluation}
To examine the performance of LiDAR odometry,
we utilize the widely-used KITTI odometry dataset~\cite{geiger2012kitii} to compare our proposed method with the state-of-the-art LiDAR-only approaches that employ different types of maps. 
Fig.~\ref{fig:traj_kitti} plots the estimated trajectories from the sequence `00' to `10', including urban, countryside, residential, and highway environments. We ran experiments for KISS-ICP and FLOAM based on their experimental setups, and the other results are directly imported from their published papers. 
For the KITTI dataset, we use the relative translational error in $\%$ and the relative rotational error in degrees per 100m for evaluation. 
   \begin{figure}[thpb]
      \centering
      \includegraphics[width=0.95\linewidth]{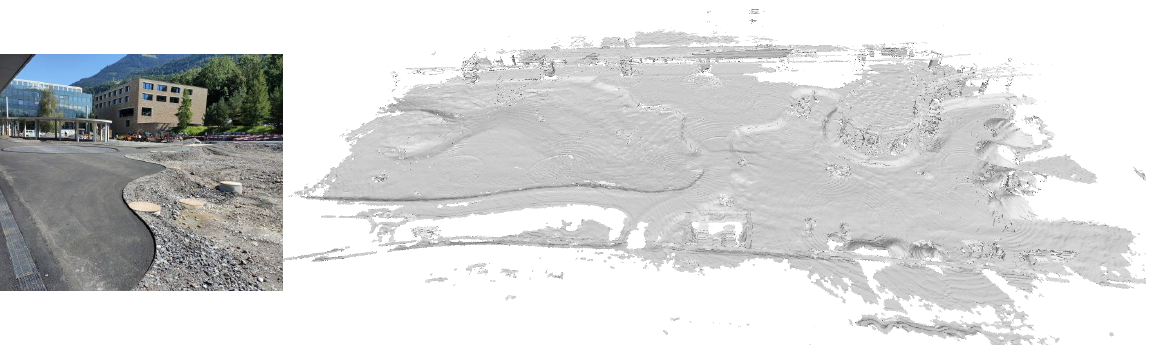}
      \caption{The left figure shows a view of the scene. The right one displays Our odometry and mapping result on the construction environment of the Hilti `cons2' sequence.
      }
      \label{fig:reconstruction_hilti}
   \end{figure}
Table~\ref{tab:kitti} shows that our approach achieves promising results with 0.51$\%$ drift in translation error and 0.15 deg/100m in rotation error in average 11 sequences, which is slightly inferior to KISS-ICP with 0.50$\%$ error in translation and better than surfel-based, NDT-based and other mesh-based methods.
Note that KISS-ICP is the state-of-the-art LiDAR-only odometry method, which performs the best among the published results. As shown in table~\ref{tab:kitti}, our results of sequences `04' `05' `06' are worse than KISS-ICP. 
This may be attributed to the 0.2m search interval used in the experiment, which hinders from finding suitable correspondences. Thus, it may prevent the convergence. Our point-to-mesh odometry outperforms other methods with various kinds of map.

   \begin{figure*}[thpb]
      \centering
      \includegraphics[width=\linewidth]{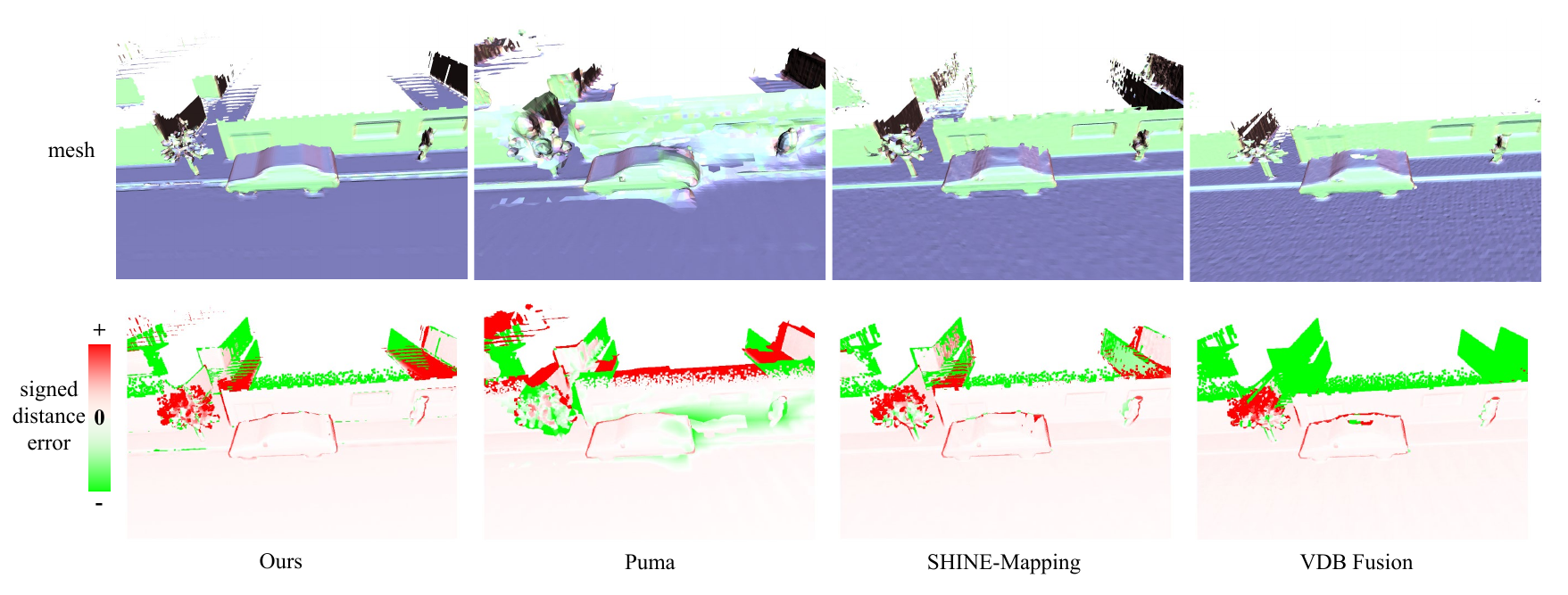}
      \caption{Qualitative comparison results on the MaiCity dataset. The first row shows the recovered mesh of different methods, including our approach, Puma~\cite{vizzo2021puma}, SHINE-Mapping~\cite{zhong2022shine}, VDB 
 Fusion~\cite{vizzo2022vdbfusion}. The second row depicts the error map between the reconstructed mesh and the ground truth. The red color represents the positive distance between the resulting mesh and the ground truth, while the green color indicates the negative distance. The brighter the color, the greater the error is.
      }
      \label{fig:colormap}
   \end{figure*}

\begin{table*}
\centering
                                                   
\caption{ Mapping results of different methods on MaiCity~\cite{vizzo2021puma} and Newer College ~\cite{ramezani2020newer} datasets in terms of map accuracy, completion, Chamfer-$L_1$ distance, completion ratio, and F-score.}
\label{tab:reconstruction}
\resizebox{\linewidth}{!}{
\begin{threeparttable}
\begin{tabular}{@{}c|ccccc|ccccc@{}}
               \toprule
               \multirow{2}{*}{Method}& \multicolumn{5}{c|}{MaiCity}& \multicolumn{5}{c}{Newer College}\\

               &Comp. $\downarrow$&Acc. $\downarrow$&C-$L_1$. $\downarrow$ &Comp.Ratio $\uparrow $ & F-score (10cm) $\uparrow $&Comp. $\downarrow$&Acc. $\downarrow$&C-$L_1$. $\downarrow$ &Comp.Ratio $\uparrow $ & F-score (20cm) $\uparrow $\\
                \midrule
                VDB Fusion~\cite{vizzo2022vdbfusion}    & 6.9          & 1.3          & 4.5          & 90.2          & 94.1 & 12.0          & 6.9          & 9.4           & 91.3                & 92.6             \\         
                Puma~\cite{vizzo2021puma}          & 32           & 1.2          & 16.9         & 78.8          & 87.3    & 15.4          & 7.7          & 11.5          & 89.9                & 91.9             \\     
                SHINE-Mapping~\cite{zhong2022shine} & 3.2          & \bf{1.1} & 2.9          & 95.2          & 95.9    & 10.0         & 6.7         & 8.4           & 93.6                & 93.7             \\      
                Ours          & \bf{2.5} & 1.2          & \bf{2.4} & \bf{96.3} & \bf{97.4} & \bf{9.6}  & \bf{6.7} & \bf{8.2}  & \bf{94.2}       & \bf{94.1}    \\
               \bottomrule
\end{tabular}

\end{threeparttable}
}
\vspace{-0.2in}
\end{table*}

\begin{table}[h]
\centering
\resizebox{\linewidth}{!}{
\begin{threeparttable}
\caption{ Evaluation of Information Dropout Ratio($\%$) on the KITTI odometry dataset (Sequence `00' to `10')
} \label{tab:dropout}
\begin{tabular}{ c|c|cccc}
\toprule
\makecell{Collision Resolving \\Strategy}   &   \makecell{Voxel Deletion \\ Scheme}               & 00      & 02      & 08      & \makecell{Avg. \\ Dropout Ratio}  \\ \midrule
\multirow{2}{*}{Linear probing} & $\times$   & 3.7     & 15.9    & 16.4    & 7.4                  \\
 & \checkmark      & 1.9e-05 & 4.8e-05 & \bf{1.9e-04} & \bf{5.3e-05}        \\
\multirow{2}{*}{Quadratic probing} &  $\times$   & 3.5     & 15.8    & 16.3    & 7.4                  \\
 & \checkmark     & 2.4e-05 & 6.7e-05 & 2.0e-04 & 5.9e-05              \\
\multirow{2}{*}{Robin hood hashing} &$\times$ & 6.64    & 17.2    & 17.6    & 8.5                  \\
 & \checkmark  & \bf{1.9e-06} & \bf{2.3e-06} & 3.0e-04 & 6.4e-05    \\ \bottomrule
\end{tabular} 
\end{threeparttable}
}
\end{table}

We quantitatively evaluate our odometry performance on the Hilti 2021 dataset~\cite{helmberger2022hilti}, which is more challenging. 
Since most sequences in the Hilti 2021 dataset provide 3-DoF ground truth poses,  we use the absolute trajectory error (ATE) in $\%$ as evaluation criteria. We only use data collected from Ouster OS0-64 liDAR for experiments. Since there are no published results for FLOAM~\cite{wang2021f} and KISS-ICP~\cite{vizzo2023kiss} on this dataset, we conducted the experiments using their implementation.
As shown in table~\ref{tab:hilit} and Fig.~\ref{fig:traj_hilti}, our proposed method achieves the best performance in most outdoor and indoor scenes on the handheld Hilti dataset 2021. Especially in the `camp2' sequence, both FLOAM and KISS-ICP perform poorly, while our method still achieves low drift in this challenging sequence.
The promising experimental results indicate that our proposed point-to-mesh odometry is robust to noise and finds reliable the correspondences.

\subsection{Mapping Evaluation}
To show the effectiveness and generalizability of our Mesh-LOAM, we qualitatively display some odometry and mapping results on two large-scale datasets~\cite{geiger2012kitii, helmberger2022hilti}.
Fig.~\ref{fig:kitti00} shows result of our proposed Mesh-LOAM with geometric details on the KITTI sequence `00', whose travel length is 3.7km. Moreover, an outdoor large-scale scene on KITTI sequence `07' is presented in Fig.~\ref{fig:reconstruction_kitti07}, where we can see smooth and flat road surfaces, as well as a variety of largely intact shaped vehicles. Furthermore, Fig.~\ref{fig:reconstruction_hilti} illustrates a real-world construction environment on the Hilti dataset, depicting smooth floors and recessed rough construction floors. Since the data is collected from a handheld platform, the point cloud coverage is not uniform, resulting in some blank edges.
These visible results demonstrate that our proposed Mesh-LOAM can generate the accurate 6-DoF poses of a mobile system and simultaneously reconstruct a dense mesh of the large-scale outdoor scene.

To visually illustrate the recovered mesh, we compare our mapping results on different grid resolutions varying from 0.1m to 0.4m. As shown in Fig.~\ref{fig:kitti07} and Fig.~\ref{fig:maicity}, our method is able to construct complete meshes for large-scale outdoor scenes while preserving accurate structures, such as the outline of vehicles, shallow curbs, and trees.
To evaluate the mapping quality of our proposed method, 
we compare our approach against three state-of-the-art methods, including a TSDF fusion-based method VDB Fusion~\cite{vizzo2022vdbfusion}, 
a Possion regression-based approach Puma~\cite{vizzo2021puma} and a learning-based method SHINE-Mapping~\cite{zhong2022shine}. We conducted experiments with the same voxel size of 0.1m on the Mai City dataset and the Newer College dataset. 
To compensate for the errors resulting from LiDAR odometry, we directly employ the ground truth pose for all methods and follow the settings of~\cite{zhong2022shine} to facilitate fair comparisons. We use five evaluation metrics~\cite{mescheder2019metrics}, including accuracy, completion, Chamfer-$L_1$ distance, completion ratio, and F-score. 

We evaluate our proposed approach on the virtual dataset Mai City and real-world dataset Newer College. 
Table~\ref{tab:reconstruction} presents the evaluation results with distance error in cm. It shows the completion Ratio and F-score, expressed as percentages, with a 10cm and 20 cm error threshold for the two datasets, respectively. 
Our proposed approach outperforms the three methods on both two datasets. 
As shown in Fig.~\ref{fig:colormap}, our method recovers the most complete surface mesh while preserving detailed structures, including vehicle outlines, pedestrian, and roadside trees. Both quantitative and qualitative results on two datasets demonstrate that our proposed meshing method is capable of recovering the complete and accurate mesh for large-scale outdoor scenes by taking advantage of our proposed hybrid-weighted voxel integration method.

\subsection{Ablation Study}
\textbf{Point-to-mesh Odometry.} To evaluate our proposed point-to-mesh odometry, we optimize the pose by minimizing the point-to-plane loss, and keep everything else exactly the same, including the same parameters. The point-to-plane loss refers to the distance from the source point cloud to the local plane in the global point cloud map. We conducted experiments on the KITTI odometry dataset. As shown in table~\ref{tab:kitti}, the KITTI relative
translation and rotation errors of point-to-plane odometry are respectively 0.71 $\%$ and 0.19 degree/100m, which are worse than our results of point-to-mesh loss. The point-to-plane loss may find some unreasonable correspondences and increase some errors. This not only shows that our point-to-mesh ICP can achieve accurate odometry accuracy but also proves the reconstructed mesh map is beneficial to our odometry.

\textbf{Voxel Deletion Scheme.}
In our implementation, open addressing~\cite{flajolet1998probing, celis1985robin} is chosen as the collision resolving strategy. To examine the efficacy of our presented voxel deletion scheme, we conduct ablation studies on the KITTI odometry dataset~\cite{geiger2012kitii} with different techniques, including robin hood hashing, linear probing, and quadratic probing. We introduce the Dropout Ratio in \% to describe the rate of lost voxels. The lower the value, the better the information is preserved.  We empirically set $2i + 1$ as linear probing and $i^2 + 1$ as quadratic probing. 

As shown in Table~\ref{tab:dropout}, all three probing methods with our voxel deletion scheme achieve a small dropout ratio on the KITTI odometry dataset, including long sequences `00', `02', and `08'. On the contrary, the results without the voxel deletion scheme perform poorly. The experiment indicates that our proposed scheme can effectively minimize the influence of hash collisions, which supports incremental reconstruction within limited memory and ensures mesh quality with minimal impact. Table~\ref{tab:dropout} also shows that linear probing $2i + 1$ with the presented strategy is appropriate for the mapping task on the KITTI odometry dataset. 

\subsection{Evaluation on Computational Efficiency}
To demonstrate the efficiency of our presented approach, we evaluate the computational time per frame on different steps, including preprocessing, point-to-mesh odometry, and incremental voxel meshing. All evaluations were conducted on the KITTI odometry dataset using a voxel size of 0.1 m. The preprocessing step requires approximately 4.7 ms per frame, while point-to-mesh odometry consumes around 11.1 ms per frame, and incremental voxel meshing takes 2.7 ms per frame.
Our method runs around 54 frames per second (fps) overall and satisfies the real-time requirement. The run-time performance is mainly due to the passive SDF computational model and the scalable partition module, which takes advantage of an efficient parallel spatial-hashing scheme. The bottleneck of speed mainly results from multiple searches for correct point-to-mesh correspondences in the point-to-mesh odometry step. With fast accessible SDF maps, we calculate the neighboring meshes during data association to speed up. However, this process still results in some time consumption.

\section{CONCLUSIONS}
This paper proposed a real-time large-scale LiDAR odometry and meshing approach. The incremental voxel meshing algorithm, utilizing a parallel spatial-hashing scheme, was introduced to quickly reconstruct triangular meshes, which integrates each LiDAR scan with only one traversal and takes advantage of a scalable partition module. Moreover, point-to-mesh odometry was designed to estimate poses between incoming point clouds and reconstructed triangular meshes.
We conducted our experiments on large-scale outdoor datasets, whose promising results demonstrated that our proposed Mesh-LOAM achieves low drift in odometry and high-quality 3D reconstruction at a fast speed.

Since mesh extraction is conducted on GPU, it requires some GPU RAM. For future work, we will explore the mesh simplification technique to reduce memory usage.

{\small
    \bibliographystyle{IEEEtran}
    \bibliography{egbib}
}

\end{document}